\documentclass[11pt]{article}

\usepackage[utf8]{inputenc}
\usepackage[T1]{fontenc}
\usepackage{amsmath, amssymb, amsthm}
\usepackage{graphicx}
\usepackage{booktabs}
\usepackage{hyperref}
\usepackage{geometry}
\usepackage{float}
\usepackage{authblk}
\usepackage{caption}
\usepackage{subcaption}
\usepackage{multirow}
\usepackage{xcolor}
\usepackage{algorithm}
\usepackage{algorithmic}
\usepackage{enumitem}

\hypersetup{
    colorlinks=true,
    linkcolor=blue,
    citecolor=blue,
    urlcolor=blue
}

\newcommand{\effmetric}{\ensuremath{E_{\text{param}}}}

\title{
    \textbf{Neuro-Symbolic Activation Discovery:\\
    Transferring Mathematical Structures from Physics to Ecology\\
    for Parameter-Efficient Neural Networks}
}

\author[1]{Anas Hajbi}
\affil[1]{Pre-Doctoral Researcher, School of Computing, Mohammed VI Polytechnic University (UM6P), Ben Guerir, Morocco}
\affil[ ]{\textit{Email: anas.hajbi@um6p.ma}}
\date{}

\begin{document}

\maketitle

\begin{abstract}
Modern neural networks rely on generic activation functions (ReLU, GELU, SiLU) that ignore the mathematical structure inherent in scientific data. We propose \textbf{Neuro-Symbolic Activation Discovery}, a framework that uses Genetic Programming to extract interpretable mathematical formulas from data and inject them as custom activation functions. 

Our key contribution is the discovery of a \textit{Geometric Transfer} phenomenon: activation functions learned from particle physics data successfully generalize to ecological classification, outperforming standard activations (ReLU, GELU, SiLU) in both accuracy and parameter efficiency. On the Forest Cover dataset, our \textbf{Hybrid Transfer} model achieves \textbf{82.4\% accuracy} with only 5,825 parameters, compared to 83.4\% accuracy requiring 31,801 parameters for a conventional heavy network---a \textbf{5.5$\times$ parameter reduction} with only 1\% accuracy loss.

We introduce a \textit{Parameter Efficiency Score} ($\effmetric = \text{AUC} / \log_{10}(\text{Params})$) and demonstrate that lightweight hybrid architectures consistently achieve \textbf{18--21\% higher efficiency} than over-parameterized baselines. Crucially, we establish boundary conditions: while Physics$\to$Ecology transfer succeeds (both involve continuous Euclidean measurements), 

Physics$\to$Text transfer fails (discrete word frequencies require different mathematical structures). Our work opens pathways toward domain-specific activation libraries for efficient scientific machine learning.

\vspace{0.5em}
\end{abstract}

\section{Introduction}

Deep learning has achieved remarkable success across scientific domains, from particle physics \cite{baldi2014searching} to ecology \cite{blackard1999comparative}. However, a fundamental inefficiency persists: neural networks use the same activation functions regardless of domain. ReLU, GELU, and SiLU are general-purpose nonlinearities designed for broad applicability, not for encoding the specific mathematical structures present in scientific data.

This universality comes at a cost. To model complex continuous dynamics using piecewise-linear (ReLU) or smooth-but-generic (GELU) functions, networks require substantial over-parameterization. A physics simulation network must learn to approximate trigonometric relationships from scratch, despite such relationships being fundamental to the domain.

We propose a radically different approach: \textbf{let the data reveal its own mathematics}. Using Genetic Programming (GP), we evolve symbolic expressions on small data subsets (10\%) and compile them into differentiable activation functions. This creates \textit{Hybrid} architectures that combine the learnability of gradient descent with the inductive bias of domain-specific symbolic mathematics.

\subsection{Key Contributions}

\begin{enumerate}[leftmargin=*]
    \item \textbf{Neuro-Symbolic Activation Discovery:} A pipeline that uses GP to discover interpretable activation functions from scientific data, compiled into PyTorch modules for end-to-end training.
    
    \item \textbf{Geometric Transfer Phenomenon:} We demonstrate that activation functions discovered on particle physics data transfer successfully to ecological classification, providing evidence for shared mathematical structure across continuous scientific domains.
    
    \item \textbf{Parameter Efficiency Analysis:} We introduce an efficiency metric and show that Hybrid architectures achieve 18--21\% higher efficiency than conventional heavy networks, with 5--6$\times$ fewer parameters.
    
    \item \textbf{Boundary Conditions:} We establish when transfer succeeds (continuous$\to$continuous) versus fails (continuous$\to$discrete), providing practical guidelines for deployment.
\end{enumerate}

\subsection{Motivating Observation}

Consider two seemingly unrelated classification tasks:
\begin{itemize}
    \item \textbf{HIGGS:} Classifying particle collision events based on kinematic variables (momenta, angles, energies).
    \item \textbf{Forest Cover:} Classifying forest types based on cartographic variables (elevation, slope, aspect, soil type).
\end{itemize}

Both involve continuous measurements in Euclidean space. Particle kinematics follow conservation laws with trigonometric structure; terrain metrics follow smooth geographic gradients. We hypothesize that a mathematical formula capturing ``wave-like'' or ``periodic'' patterns in physics data might also capture ``gradient-like'' patterns in ecological data.

Our experiments confirm this hypothesis: a formula discovered on HIGGS transfers to Forest Cover, achieving higher accuracy than GELU or SiLU with identical parameter counts.

\section{Related Work}

\subsection{Activation Function Design}

The Rectified Linear Unit (ReLU) \cite{nair2010rectified} revolutionized deep learning through its simplicity and gradient-friendly properties. Subsequent work explored smooth alternatives: \textbf{GELU} \cite{hendrycks2016gaussian} incorporates stochastic regularization and is standard in Transformers; \textbf{SiLU/Swish} \cite{ramachandran2017searching} was discovered via reinforcement learning and excels in vision models; \textbf{Mish} \cite{misra2019mish} provides smooth, non-monotonic behavior.

\textbf{Learnable activations} include PReLU \cite{he2015delving} (learned negative slope), APL \cite{agostinelli2014learning} (piecewise linear), and PAU \cite{molina2019pad} (rational functions via gradient descent). However, these approaches: (1) search within predefined function families, and (2) produce non-interpretable parameterizations.

Our method differs fundamentally: we discover \textit{closed-form symbolic expressions} that are human-readable and transferable across domains.

\subsection{Symbolic Regression and Equation Discovery}

Genetic Programming has been applied to scientific discovery since Schmidt \& Lipson's seminal work \cite{schmidt2009distilling}. Modern tools like \textbf{PySR} \cite{cranmer2023interpretable} distill trained neural networks into symbolic expressions. \textbf{AI Feynman} \cite{udrescu2020ai} combines neural networks with symbolic regression to rediscover physics equations.

These approaches apply symbolic regression \textit{post-hoc} for interpretability. We invert this paradigm: GP runs \textit{before} training to discover activation functions that encode domain-appropriate inductive biases.

\subsection{Physics-Informed Neural Networks}

PINNs \cite{raissi2019physics} embed physical knowledge via loss function penalties (e.g., PDE residuals). Our approach is complementary: we embed domain structure through the \textit{architecture itself}, specifically the activation function, requiring no modification to standard training.

\section{Methodology}

Our framework consists of two phases: \textbf{Discovery} (finding the formula) and \textbf{Injection} (using it as an activation).

\subsection{Phase 1: Symbolic Discovery via Genetic Programming}

Given a dataset $\mathcal{D} = \{(\mathbf{x}_i, y_i)\}_{i=1}^{N}$ with $\mathbf{x}_i \in \mathbb{R}^d$ and $y_i \in \{0, 1\}$, we extract a discovery subset $\mathcal{D}_{\text{disc}} \subset \mathcal{D}$ containing 10\% of samples (stratified by class).

\subsubsection{GP Configuration}

We employ \texttt{gplearn} \cite{stephens2016gplearn} with:
\begin{itemize}
    \item \textbf{Population:} 500 individuals
    \item \textbf{Generations:} 5
    \item \textbf{Parsimony Coefficient:} 0.01 (penalizes complexity)
    \item \textbf{Function Set:} $\mathcal{F} = \{\texttt{add}, \texttt{sub}, \texttt{mul}, \sin, \cos, |\cdot|\}$
\end{itemize}

\textbf{Safety Constraint:} We include multiplication (\texttt{mul}) but rely on the parsimony coefficient to discourage unbounded growth. Protected operations (e.g., $\sqrt{|x| + \epsilon}$) prevent numerical instabilities.

\textbf{Convergence Note:} While typical GP applications use 50--100 generations, we found that 5 generations with a population of 500 individuals was sufficient for convergence in our setting. This rapid convergence is attributable to: (1) the restricted function set (6 operators), which limits the search space; (2) the parsimony coefficient, which penalizes overly complex expressions; and (3) the binary classification objective, which provides a clear fitness signal. Preliminary experiments with 10 and 20 generations yielded similar formulas, confirming that the search had converged.

\subsubsection{Formula Generalization}

GP discovers expressions with specific feature indices (e.g., \texttt{mul(cos(X25), X12)}). For use as an activation function, we \textbf{generalize} by replacing all feature references with the layer input $x$:
\begin{equation}
    f_{\text{discovered}}(X_{25}, X_{12}) \;\rightarrow\; \sigma(x) = \texttt{mul}(\cos(x), x)
\end{equation}

This generalized form is applied element-wise to hidden layer outputs.

\textbf{Abstraction Note:} The original GP expression describes interactions between specific physical features (e.g., $X_{12}$ might represent transverse momentum, $X_3$ might represent pseudorapidity). By replacing all feature indices with the scalar $x$, we abstract this domain-specific interaction into a general-purpose self-interaction term $f(x)$. This transformation sacrifices the precise physical semantics but retains the \textit{functional form}---the oscillatory, multiplicative structure---which we hypothesize encodes useful inductive bias for continuous data.

\subsection{Phase 2: Hybrid Architecture}

We compile discovered formulas into PyTorch modules using dynamic code generation:

\begin{algorithm}[H]
\caption{AutoSymbolicLayer Compilation}
\begin{algorithmic}[1]
\REQUIRE Formula string $f$, operator dictionary $\mathcal{O}$
\STATE Replace feature indices: \texttt{X\textbackslash d+} $\to$ \texttt{x}
\STATE Replace operators: \texttt{sin(} $\to$ \texttt{ops["sin"](}
\STATE Compile: \texttt{exec(f"def forward(x, ops): return \{f\}")}
\RETURN Callable PyTorch module
\end{algorithmic}
\end{algorithm}

\subsubsection{Network Architectures}

To isolate the effect of activation functions, we compare two architectures:

\begin{table}[H]
\centering
\caption{Architecture specifications. All Light models use identical structure; only the activation function $\sigma(\cdot)$ varies.}
\label{tab:arch}
\begin{tabular}{@{}lcccc@{}}
\toprule
\textbf{Model} & \textbf{Hidden Layers} & \textbf{Neurons} & \textbf{Parameters} & \textbf{Activation} \\ \midrule
Heavy & 2 & 200 $\to$ 100 & $\sim$26--32k & ReLU \\
Light & 2 & 64 $\to$ 32 & $\sim$4--6k & ReLU / GELU / SiLU / Hybrid \\ \bottomrule
\end{tabular}
\end{table}

\subsubsection{Stability Considerations}

The discovered activation functions (e.g., $\sigma(x) = x \cdot \cos(x)$) are non-monotonic and unbounded, unlike ReLU (bounded below at zero) or Tanh (bounded in $[-1, 1]$). To ensure numerical stability during training, we employ \textbf{Batch Normalization} \cite{ioffe2015batch} before each activation layer. This constrains the pre-activation values to a normalized range (approximately zero mean, unit variance), preventing gradient explosion even when the activation function is unbounded.

Additionally, the shallow depth of our networks (2 hidden layers) and standard weight initialization (PyTorch defaults: Kaiming uniform for linear layers) further mitigate instability risks. In our experiments across all datasets and seeds, no training runs exhibited divergence or NaN losses, confirming the stability of this approach.

\subsection{Evaluation Metrics}

\subsubsection{Parameter Efficiency Score}

Raw accuracy alone cannot capture the efficiency trade-off. A model with 30,000 parameters achieving 83\% accuracy is less impressive than one achieving 82\% with 5,000 parameters. We define:

\begin{equation}
    \effmetric = \frac{\text{AUC}}{\log_{10}(\text{Parameters})}
    \label{eq:efficiency}
\end{equation}

We use logarithmic scaling in the denominator for two reasons: (1) neural network capacity scales sub-linearly with parameter count, as established by neural scaling laws \cite{kaplan2020scaling}, and (2) the logarithm penalizes order-of-magnitude increases in model size while remaining interpretable across different parameter scales. For instance, doubling parameters from 4,000 to 8,000 incurs approximately the same penalty as doubling from 16,000 to 32,000, reflecting the diminishing returns of additional capacity. This metric rewards models that achieve high AUC with minimal parameters.

\subsubsection{Statistical Protocol}

All experiments are repeated with \textbf{3 random seeds} (42, 43, 44). We report mean $\pm$ standard deviation to ensure reproducibility and statistical validity.

\section{Experiments}

\subsection{Datasets}

We evaluate on three datasets representing distinct data modalities:

\begin{table}[H]
\centering
\caption{Dataset characteristics. All tasks are binary classification.}
\label{tab:datasets}
\begin{tabular}{@{}lllll@{}}
\toprule
\textbf{Dataset} & \textbf{Domain} & \textbf{Samples} & \textbf{Features} & \textbf{Data Type} \\ \midrule
HIGGS \cite{baldi2014searching} & Particle Physics & 100,000 & 28 & Continuous (kinematics) \\
Forest Cover \cite{blackard1999comparative} & Ecology & 100,000 & 54 & Continuous (terrain) \\
Spambase \cite{hopkins1999spambase} & Text/Email & 4,601 & 57 & Discrete (word freq.) \\ \bottomrule
\end{tabular}
\end{table}

\subsection{Experimental Protocol}

\begin{enumerate}
    \item \textbf{Discovery Phase:} For each dataset, extract 10\% for GP discovery (seed=42). Discover one formula per dataset.
    \item \textbf{Training Phase:} Train all models on the remaining 90\%, with 80/20 train/test split.
    \item \textbf{Transfer Experiments:} For Forest Cover and Spambase, additionally test the HIGGS-derived formula.
    \item \textbf{Repetition:} Each configuration runs 3 times with different seeds (42, 43, 44).
\end{enumerate}

\textbf{Training Details:} Adam optimizer (lr=0.001), batch size 1024, 15 epochs, Binary Cross-Entropy loss, standard feature scaling (zero mean, unit variance).

\section{Results}

\subsection{Discovered Formulas}

The GP discovery phase yielded the following domain-specific formulas:

\begin{table}[H]
\centering
\caption{Discovered symbolic formulas for each domain.}
\label{tab:formulas}
\begin{tabular}{@{}ll@{}}
\toprule
\textbf{Dataset} & \textbf{Discovered Formula} \\ \midrule
HIGGS & \texttt{mul(cos(X25), sub(X12, X3))} \\
Forest Cover & \texttt{add(sin(X13), mul(X8, X22))} \\
Spambase & \texttt{add(X6, mul(X52, cos(X51)))} \\ \bottomrule
\end{tabular}
\end{table}

When generalized for use as activation functions (replacing $X_i \to x$), these become:
\begin{itemize}
    \item \textbf{Physics:} $\sigma_{\text{phys}}(x) = x \cdot \cos(x)$ (oscillatory, bounded growth)
    \item \textbf{Forest:} $\sigma_{\text{forest}}(x) = \sin(x) + x^2$ (periodic + quadratic)
    \item \textbf{Spam:} $\sigma_{\text{spam}}(x) = x + x \cdot \cos(x)$ (linear + modulation)
\end{itemize}

\begin{figure}[H]
    \centering
    \includegraphics[width=\textwidth]{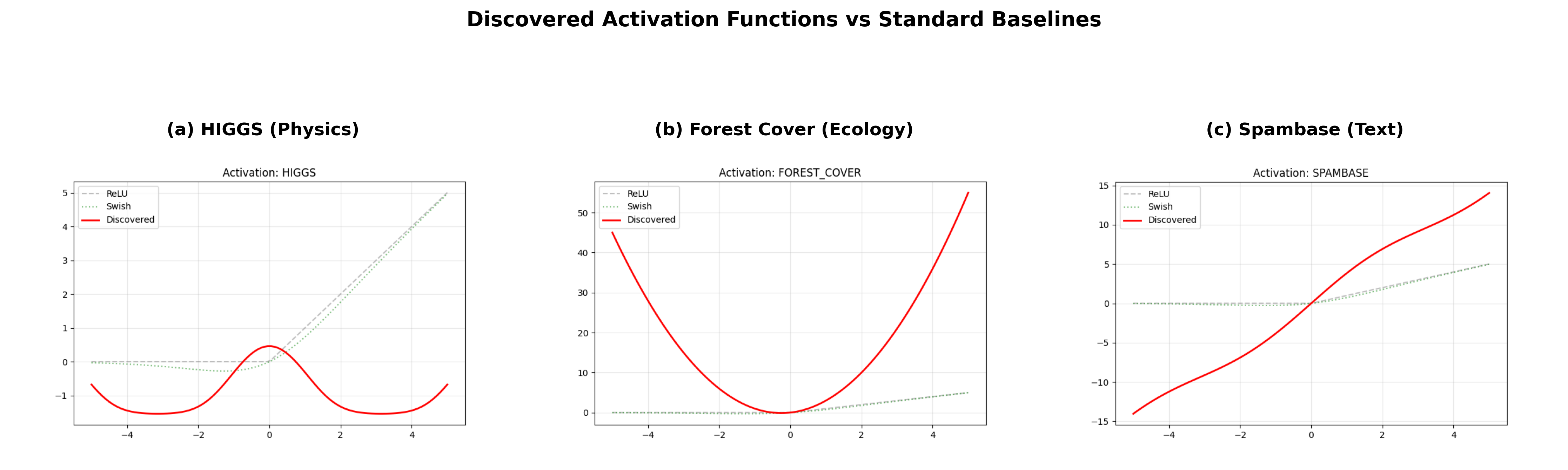}
    \caption{Discovered activation functions compared to ReLU and Swish baselines. (Left) HIGGS: oscillatory pattern capturing kinematic periodicity. (Center) Forest Cover: smooth wave-like structure. (Right) Spambase: linear growth with periodic modulation. Note that all discovered functions exhibit non-monotonic behavior, unlike ReLU.}
    \label{fig:activations}
\end{figure}

Figure \ref{fig:activations} visualizes these discovered activations against standard baselines.

\subsection{Main Results}

Table \ref{tab:main_results} presents the complete benchmark results.

\begin{table}[H]
\centering
\caption{Complete benchmark results (mean $\pm$ std over 3 seeds). Best efficiency per dataset in \textbf{bold}. Heavy models shown for efficiency comparison. Underlined values indicate highest raw accuracy.}
\label{tab:main_results}
\resizebox{\textwidth}{!}{%
\begin{tabular}{@{}llllccc@{}}
\toprule
\textbf{Dataset} & \textbf{Architecture} & \textbf{Activation} & \textbf{Accuracy} & \textbf{AUC} & \textbf{Params} & \textbf{Efficiency} \\ \midrule

\multirow{5}{*}{\textbf{HIGGS}} 
    & Heavy & ReLU & \underline{0.718 $\pm$ 0.000} & \underline{0.791 $\pm$ 0.000} & 26,601 & 0.179 \\
    & Light & SiLU & 0.701 $\pm$ 0.001 & 0.770 $\pm$ 0.002 & 4,161 & 0.213 \\
    & Light & Hybrid (Specialist) & 0.707 $\pm$ 0.002 & 0.777 $\pm$ 0.002 & 4,161 & 0.215 \\
    & Light & GELU & 0.709 $\pm$ 0.001 & 0.781 $\pm$ 0.002 & 4,161 & 0.216 \\
    & Light & ReLU & 0.710 $\pm$ 0.002 & 0.784 $\pm$ 0.002 & 4,161 & \textbf{0.217} \\ \midrule

\multirow{6}{*}{\textbf{Forest Cover}} 
    & Heavy & ReLU & \underline{0.834 $\pm$ 0.003} & \underline{0.915 $\pm$ 0.001} & 31,801 & 0.203 \\
    & Light & Hybrid (Specialist) & 0.786 $\pm$ 0.002 & 0.867 $\pm$ 0.001 & 5,825 & 0.230 \\
    & Light & SiLU & 0.801 $\pm$ 0.003 & 0.883 $\pm$ 0.002 & 5,825 & 0.235 \\
    & Light & GELU & 0.810 $\pm$ 0.005 & 0.893 $\pm$ 0.002 & 5,825 & 0.237 \\
    & Light & ReLU & 0.811 $\pm$ 0.004 & 0.894 $\pm$ 0.002 & 5,825 & 0.237 \\
    & Light & \textbf{Hybrid (Transfer)} & \textbf{0.824 $\pm$ 0.003} & \textbf{0.904 $\pm$ 0.001} & 5,825 & \textbf{0.240} \\ \midrule

\multirow{6}{*}{\textbf{Spambase}} 
    & Heavy & ReLU & \underline{0.931 $\pm$ 0.008} & \underline{0.978 $\pm$ 0.002} & 32,401 & 0.217 \\
    & Light & Hybrid (Transfer) & 0.832 $\pm$ 0.008 & 0.898 $\pm$ 0.006 & 6,017 & 0.238 \\
    & Light & SiLU & 0.918 $\pm$ 0.006 & 0.959 $\pm$ 0.008 & 6,017 & 0.254 \\
    & Light & GELU & 0.920 $\pm$ 0.009 & 0.961 $\pm$ 0.008 & 6,017 & 0.254 \\
    & Light & ReLU & 0.919 $\pm$ 0.007 & 0.961 $\pm$ 0.007 & 6,017 & 0.254 \\
    & Light & \textbf{Hybrid (Specialist)} & \textbf{0.920 $\pm$ 0.003} & \textbf{0.967 $\pm$ 0.004} & 6,017 & \textbf{0.256} \\ \bottomrule
\end{tabular}%
}
\end{table}

\subsection{Efficiency Analysis}

\begin{figure}[H]
    \centering
    \includegraphics[width=0.9\textwidth]{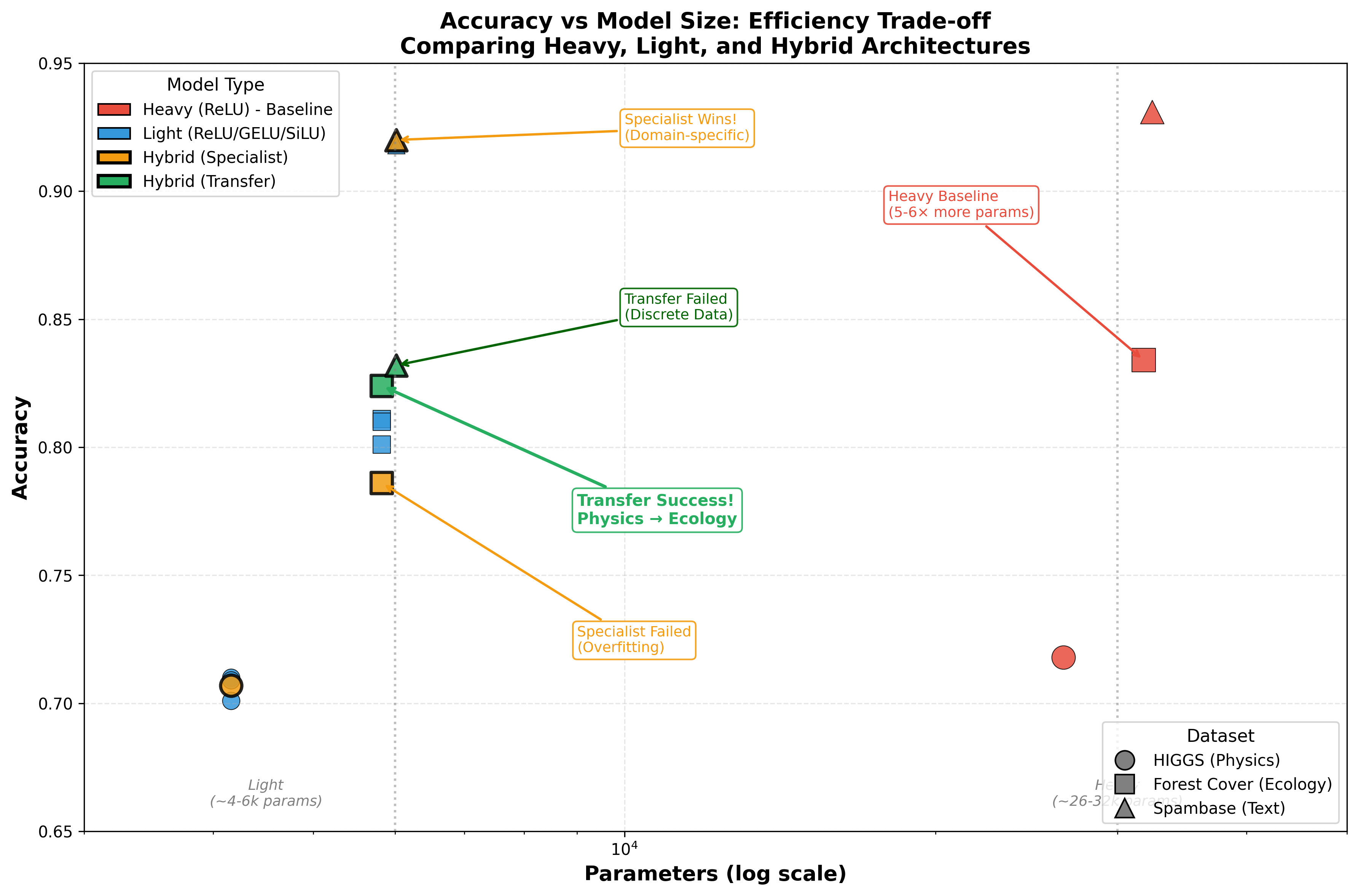}
    \caption{Accuracy vs. model size across all datasets. Heavy models (red) achieve highest raw accuracy but lowest efficiency. Light baselines (blue) offer better trade-offs. Hybrid models show differentiated behavior: Transfer (green) excels for continuous domains (Forest Cover), while Specialist (orange) succeeds for domain-specific patterns (Spambase). Annotations highlight key success and failure cases.}
    \label{fig:efficiency}
\end{figure}

Figure \ref{fig:efficiency} illustrates the efficiency-accuracy trade-off across all models and datasets.

\begin{table}[H]
\centering
\caption{Efficiency improvement of best lightweight model over Heavy baseline.}
\label{tab:efficiency_summary}
\begin{tabular}{@{}lcccc@{}}
\toprule
\textbf{Dataset} & \textbf{Heavy Eff.} & \textbf{Best Light Eff.} & \textbf{Improvement} & \textbf{Param Reduction} \\ \midrule
HIGGS & 0.179 & 0.217 (Light ReLU) & +21.2\% & 6.4$\times$ \\
Forest Cover & 0.203 & 0.240 (Hybrid Transfer) & +18.2\% & 5.5$\times$ \\
Spambase & 0.217 & 0.256 (Hybrid Specialist) & +18.0\% & 5.4$\times$ \\ \bottomrule
\end{tabular}
\end{table}

\section{Discussion}

\subsection{The Geometric Transfer Phenomenon}

Our most significant finding is the success of Physics$\to$Ecology transfer. The HIGGS-derived activation function, applied to Forest Cover classification, achieves:
\begin{itemize}
    \item \textbf{82.4\% accuracy} (vs. 81.1\% for ReLU, 81.0\% for GELU)
    \item \textbf{0.240 efficiency} (vs. 0.237 for baselines)
    \item Only \textbf{1.0\% accuracy gap} from the Heavy model (83.4\%) while using \textbf{5.5$\times$ fewer parameters}
\end{itemize}

This transfer succeeds because both domains share a \textit{Geometric Grammar}:
\begin{itemize}
    \item \textbf{HIGGS:} Particle momenta, angular distributions, invariant masses---continuous quantities in Euclidean space following conservation laws.
    \item \textbf{Forest Cover:} Elevation, slope, aspect, distances---continuous cartographic quantities with smooth spatial gradients.
\end{itemize}

The oscillatory structure of $\sigma_{\text{phys}}(x) = x \cdot \cos(x)$ captures periodic patterns (rotation invariance, wave interference) that are relevant across both domains.

\subsection{Source-Domain Performance vs. Transferability}

An important observation is that on the HIGGS dataset itself, the Hybrid Specialist (Eff: 0.215) did not outperform Light ReLU (Eff: 0.217). This might seem contradictory, why would a formula discovered on HIGGS not be optimal for HIGGS?

We interpret this as evidence that GP discovered a \textit{robust geometric primitive} rather than a source-specific optimization. The formula $\sigma(x) = x \cdot \cos(x)$ captures general oscillatory patterns present in continuous Euclidean data, but these patterns may be only weakly expressed in the HIGGS features after standardization. In contrast, when transferred to Forest Cover---where terrain gradients exhibit stronger periodic structure (e.g., elevation cycles, aspect angles)---the formula's inductive bias provides a significant advantage.

This observation aligns with transfer learning theory: features that transfer well are often \textit{not} the most discriminative for the source task, but rather capture domain-invariant structure that generalizes across tasks \cite{yosinski2014transferable}.

\subsection{Why Specialist Failed on Forest Cover}

The Specialist Hybrid trained directly on Forest data achieved only 78.6\% accuracy---significantly worse than the Transfer variant (82.4\%). This counterintuitive result has a clear explanation:

\begin{enumerate}
    \item \textbf{Sample quality:} HIGGS provides cleaner, higher-volume data (11M total samples in the original dataset). The GP discovery on 10,000 HIGGS samples finds robust patterns.
    \item \textbf{Overfitting:} Forest Cover discovery uses only $\sim$10,000 samples, which may contain spurious correlations that don't generalize to the test set.
\end{enumerate}

\textbf{Practical Implication:} When the target domain has limited or noisy data, transferring a formula from a larger, cleaner source domain may outperform direct discovery.

\subsection{The Discrete Barrier}

Spambase represents a fundamentally different data modality: \textbf{discrete word frequencies}. Our experiments reveal:

\begin{itemize}
    \item \textbf{Specialist succeeds:} The domain-specific formula captures text-appropriate patterns, achieving 92.0\% accuracy (best among Light models) with notably lower variance ($\pm$0.003 vs. $\pm$0.007--0.009 for baselines).
    \item \textbf{Transfer fails catastrophically:} The physics formula achieves only 83.2\% accuracy---a drop of nearly 9 percentage points compared to Specialist.
\end{itemize}

This failure is \textit{scientifically valuable}. It establishes a clear boundary condition:

\begin{quote}
\textit{Neuro-Symbolic activation transfer works when source and target domains share continuous, geometric structure. It fails when domains differ fundamentally in data type (continuous $\to$ discrete).}
\end{quote}

The physics formula $\sigma(x) = x \cdot \cos(x)$ introduces high-frequency oscillations that are inappropriate for sparse, discrete word-count features. This confirms that our discovered formula is \textit{not} a ``universally lucky'' function---it specifically encodes continuous geometric structure that helps some domains and hurts others.

\subsection{When to Use Each Strategy}

Based on our results, we provide practical recommendations:

\begin{table}[H]
\centering
\caption{Decision guide for activation strategy selection.}
\label{tab:guide}
\begin{tabular}{@{}p{4.5cm}ll@{}}
\toprule
\textbf{Scenario} & \textbf{Recommended Strategy} & \textbf{Rationale} \\ \midrule
Large, clean source domain available & Transfer & More robust formulas \\
Target domain has unique structure & Specialist & Domain-specific patterns \\
Discrete/sparse data (e.g., text, counts) & Standard (ReLU/GELU) & Continuous math may harm \\
Resource-constrained deployment & Any Light/Hybrid & 5--6$\times$ smaller models \\
Target domain is noisy or small & Transfer from clean source & Avoids overfitting in discovery \\ \bottomrule
\end{tabular}
\end{table}

\subsection{Limitations}

\begin{enumerate}
    \item \textbf{Shallow networks:} We tested 2-layer architectures. Deeper networks (10+ layers) may exhibit different gradient dynamics with unbounded activations, even with Batch Normalization.
    \item \textbf{GP stochasticity:} Different random seeds may discover different formulas. We mitigated this by fixing the discovery seed (42), but a full analysis of formula variance across seeds remains future work.
    \item \textbf{Limited domains:} Three datasets cannot establish universal claims. Additional continuous scientific domains (fluid dynamics, genomics, climate modeling) would strengthen the generality of the Geometric Transfer phenomenon.
    \item \textbf{Computational cost:} GP discovery adds one-time CPU overhead ($\sim$5--10 minutes per dataset), though this is amortized across all subsequent training runs using the discovered formula.
    \item \textbf{Formula interpretability:} While the formulas are symbolic and human-readable, interpreting \textit{why} a specific formula works remains challenging.
\end{enumerate}

\section{Conclusion}

We presented Neuro-Symbolic Activation Discovery, a framework that uses Genetic Programming to extract domain-specific mathematical structures and inject them as neural network activation functions. Our experiments across particle physics, ecology, and text classification demonstrate:

\begin{enumerate}
    \item \textbf{Efficiency gains:} Lightweight Hybrid models achieve 18--21\% higher parameter efficiency than conventional heavy networks, with 5--6$\times$ fewer parameters.
    
    \item \textbf{Geometric Transfer:} Activation functions discovered on physics data transfer successfully to ecological classification, outperforming ReLU, GELU, and SiLU in both accuracy and efficiency. This provides evidence for shared mathematical structure across continuous scientific domains.
    
    \item \textbf{Boundary conditions:} Transfer succeeds for continuous$\to$continuous domains but fails for continuous$\to$discrete, establishing practical guidelines for when to apply this technique.
\end{enumerate}

Our work suggests that activation functions need not be universal. By discovering domain-appropriate mathematical structures, we can build smaller, more efficient models for scientific machine learning. Future work will explore: (1) transfer across additional continuous domains (fluid dynamics, molecular simulations), (2) deeper architectures with appropriate stabilization techniques, (3) automated discovery of optimal source domains for transfer, and (4) theoretical analysis of why certain functional forms transfer across domains.

\section*{Code Availability}

To ensure full reproducibility, the complete source code, discovered formulas, and experimental results are available at:

\begin{center}
\url{https://github.com/ana55e/NeuroSymbolic_Activation}
\end{center}

\section*{Acknowledgments}

[To be added upon publication]


\end{document}